\pdfoutput=1
\documentclass[11pt]{article}

\usepackage{acl}

\usepackage{times}
\usepackage{latexsym}
\usepackage{graphicx} %插入图片的宏包
\usepackage{float} %设置图片浮动位置的宏包
\usepackage{subfigure} %插入多图时用子图显示的宏包
\usepackage{algorithm}
\usepackage{algorithmic}
\usepackage{amsmath}
\usepackage{booktabs}

\usepackage[T1]{fontenc}
\usepackage[utf8]{inputenc}

\usepackage{microtype}

\title{A Densely Connected Criss-Cross Attention Network for Document-level Relation Extraction}

\author {
    Liang Zhang\textsuperscript{\rm 1} 
    Yidong Cheng \textsuperscript{\rm 1}\\
    \textsuperscript{\rm 1} Department of Artificial Intelligence, School of Informatics, Xiamen University\\
    \texttt{ \{lzhang,ydcheng\}@stu.xmu.edu.cn} \\
}
\begin{document}
\maketitle

\begin{abstract}
Document-level relation extraction (RE) aims to identify relations between two entities in a given document. Compared with its sentence-level counterpart, document-level RE requires complex reasoning. Previous research normally completed reasoning through information propagation on the mention-level or entity-level document-graph, but rarely considered reasoning at the entity-pair-level.
% and thus ignored the correlations among the relationships.
% Moreover, the latest state-of-the-art model, DocuNet, shows that the global interdependence among triples can improve the performance of document-level RE. 
In this paper, we propose a novel model, called \textbf{Dense}ly Connected \textbf{C}riss-\textbf{C}ross Attention \textbf{Net}work (\textbf{Dense-CCNet}), for document-level RE, which can complete logical reasoning at the entity-pair-level. Specifically, the Dense-CCNet performs entity-pair-level logical reasoning through the Criss-Cross Attention (\textbf{CCA}), which can collect contextual information in horizontal and vertical directions on the entity-pair matrix to enhance the corresponding entity-pair representation. 
In addition, we densely connect multiple layers of the CCA to simultaneously capture the features of single-hop and multi-hop logical reasoning.
We evaluate our Dense-CCNet model on three public document-level RE datasets, DocRED, CDR, and GDA. Experimental results demonstrate that our model achieves state-of-the-art performance on these three datasets. 
% Furthermore, further analysis demonstrates that our model has excellent robustness in low-resource scenarios.
\end{abstract}

\section{Introduction}
Relation extraction (RE) aims to identify relationships between two entities from raw texts. It is of great importance to many real-world applications such as knowledge base construction, question answering, and biomedical text analysis \cite{c:101}. Most of the existing work focuses on sentence-level RE, which predicts the relationship between entities in a single sentence \cite{c:102,c:103}. 
However, large amounts of relationships are expressed by multiple sentences in real life \cite{c:104}. According to the statistics of the DocRED \cite{c:104} dataset which is obtained from Wikipedia documents, at least 40.7\% of relations can only be extracted from multiple sentences. 
%Therefore, document-level RE models must have the ability to extract relational facts across sentences. 
Therefore, researches on document-level RE models that can extract relational facts across sentences have gained increasing attention recently.

\begin{figure}[t]
\centering
\includegraphics[width=0.95 \columnwidth]{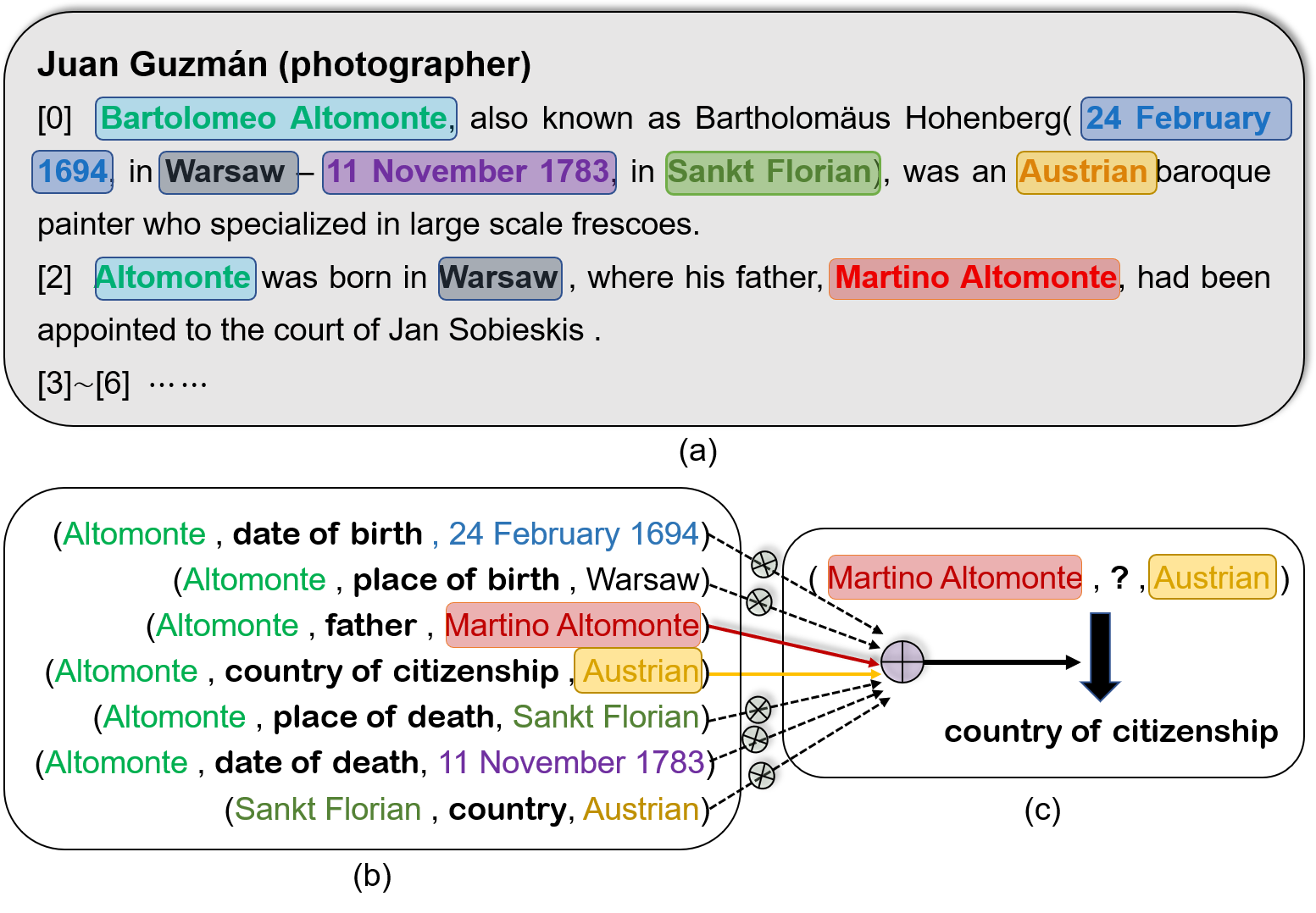} % Reduce the figure size so that it is slightly narrower than the column. Don't use precise values for figure width.This setup will avoid overfull boxes.
\caption{An example comes from the DocRED dataset, which shows that triples with overlapping entities provide important information for reasoning the complex inter-sentential relations. (a) is a document, in which different colors represent different entities. (b) lists some intra-sentential triplets, which can be easily identified. (c) shows a triple whose relationship requires logical reasoning techniques to be recognized. The arrows between (b) and (c) indicate the correlation among triples.}
\label{fig1}
\end{figure}

Compared with sentence-level RE, the main challenge is that many relations in document-level RE could only be extracted through the technique of reasoning.
Since these relationship facts are not explicitly expressed in the document, the model must captures the correlation between the relationships to infer these relationships.
Therefore, capturing the relevance of  the relationships is essential to improve the reasoning ability of document-level RE models.
% there are several main challenges in document-level RE. First, two entities involved in a relationship may appear in different sentences. Therefore, this kind of relational facts cannot be extracted by a single sentence. Second, many relations in document-level RE could only be extracted through the technique of reasoning. Third, it is very important for the document-level RE models to capture the interdependence among triples.
Figure~\ref{fig1} shows an example from the DocRED dataset. The Figure 1b lists the intra-sentential triplets, such as (\textit{Altomonte, date of birth, 24 February 1694}), (\textit{Altomonte, father, Martino Altomonte}), and (\textit{Altomonte, country of citizenship, Austrian}), which could be easily recognized since two related entities appear in the same sentence. 
However, it is non-trivial to predict the inter-sentential relations between \textit{Martino Altomonte} and \textit{Austrian} because the document does not explicitly express the relationship between them. 
In fact, the model needs to firstly capture the correlation among (\textit{Altomonte, father, Martino Altomonte}), (\textit{Altomonte, country of citizenship, Austrian}), and (\textit{Martino Altomonte, country of citizenship, Austrian}) and use logical reasoning techniques to identify this complex relationship, as shown in Figure 1c.

To extract these complex relationships, most current approaches constructed a document-level graph based on heuristics, structured attention, or dependency structures \cite{c:105,c:106,c:107,c:108}, and then perform inference with graph convolutional network (GCN) \cite{c:109,c:110} on the graph. Meanwhile, considering the transformer architecture can implicitly model long-distance dependencies, some studies \cite{c:111,c:112} directly applied pre-trained models rather than explicit graph reasoning \cite{c:113}. 
These methods captures the correlation between relationships through the information transfer between tokens, mentions or entities, which can be indirect and inefficient.

% However, these methods have two potential problems: (1) They  mainly focus on token-level syntactic features or contextual information, but ignore the interdependence among triples. (2) They perform logical reasoning only on mentions or entities level rather than entity-pairs level. To capture the interdependency among triples, DocuNet \cite{c:113} formulates document-level RE as a semantic segmentation task and captures global interdependency among triples through the U-shaped segmentation module \cite{c:115}. However, DocuNet does not consider explicit logical reasoning. 
In this paper, we use the information transfer between the entity-pairs to capture the correlation between relationships more efficiently and directly.
Moreover, as it can be seen in Figure 1, only (\textit{Altomonte, father, Martino Altomonte}) and (\textit{Altomonte, country of citizenship, Austrian}) triples, rather than the other triples, provide important information to infer the relations between \textit{Martino Altomonte} and \textit{Austrian}. 
Inspired by this phenomenon, we guess that the interaction between the triples with overlapping entities is a reasonable way of entity-pair-level reasoning.

Therefore, we propose a novel Dense-CCNet model by integrating the Criss-Cross Attention (\textbf{CCA}) \cite{c:116} into the densely connected framework \cite{c:121}. 
The CCNet model \cite{c:116} is an advanced semantic segmentation model recently proposed in the field of computer vision, which captures global context information from full-image through the CCA (as shown in Figure~\ref{fig2}). 
The CCA applied to the entity-pair matrix can realize the interaction between entity-pairs with overlapping entities, which can complete the logical reasoning of the entity-pair-level. 
To fully capture the features of single-hop and multi-hop reasoning, we stack the multi-layer modules CCA modules by the densely connected framework.
The lower layers in Dense-CCNet can capture local interdependence among entity-pairs and complete single-hop logical reasoning, while the upper layers can capture global interdependence among entity-pairs and complete multi-hop logical reasoning.

Since the CCA can only complete the reasoning mode of \textbf{A$\rightarrow{*}\rightarrow$B}, we expand the field (which single-layer CCA can pay attention to) to cover a wider range of reasoning modes, such as  \textbf{A$\rightarrow{*}\leftarrow$B},  \textbf{A$\leftarrow{*}\leftarrow$B}, and \textbf{A$\leftarrow{*}\rightarrow$B}. 
In addition, we found that more than 90\% of the entity pairs are irrelevant (that is, there is no relationship between two entities) in the document, and these entity pairs may limit the model’s reasoning ability. To reduce the influence of unrelated entity-pairs, we use two techniques:
\textbf{(1) Clustering loss}: The clustering loss separates the related entity-pairs (that is, there is relationship between two entities) from the unrelated entity-pairs in the representation space.
\textbf{(2) Attention bias}: We add a bias term to a bias term to the attention score of the CCA, which makes the CCA pay more attention to related entity pairs.

% In addition, each entity pair can harvest full entity-pair matrix contextual information from all entity-pair to generate new features with rich contextual information through the stack of two layers of cross-attention modules, and we stack cross-attention modules by the densely connected framework. The dense connection framework can reuse the low-level features of the model, so Dense-CCNet can capture the global and local interdependence at the same time. 

In summary, our main contributions are as follows:
\begin{itemize}
\item We introduce the Dense-CCNet module that can more directly and effectively model the correlation between relationships through the entity-pair-level reasoning.
\item We introduce four methods to further improve the reasoning ability of the CCA: Dense connection, Expanding the Field of Attention, Clustering loss, and Attention bias.
\item Experimental results on three public document-level RE datasets shows that our Dense-CCNet model can achieve state-of-the-art performance.
\end{itemize}

% \begin{figure}[t]
% \centering
% \includegraphics[width=0.6 \columnwidth]{LaTeX/figure2.png} % Reduce the figure size so that it is slightly narrower than the column. Don't use precise values for figure width.This setup will avoid overfull boxes.
% \caption{Illustration of criss-cross attention module from CCNet. H can be an image or entity-pair matrix.}
% \label{fig2}
% \end{figure}

\section{Methodology}
\begin{figure*}[t]
\centering
\includegraphics[width=0.9 \textwidth]{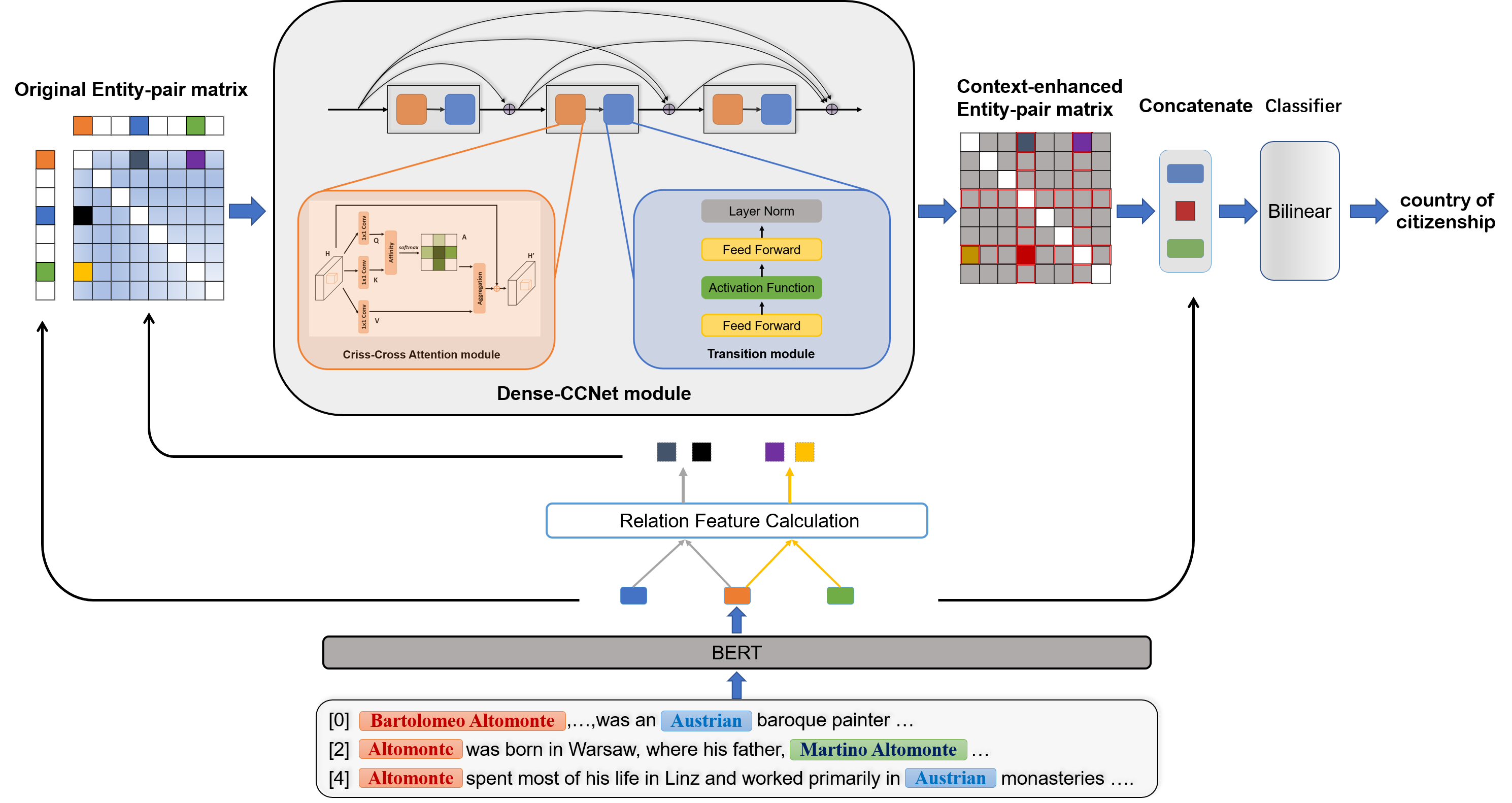} % Reduce the figure size so that it is slightly narrower than the column.
\caption{The overall architecture of our Dense-CCNet-based document-level RE model. Firstly, the BERT model encodes the input document to obtain the context embedding of each words, and then we obtains the representations of the entities ($h_{e_s}$,$h_{e_o}$) through a pooling operation. Secondly, the relation features ($M_{s,o}/M_{o,s}$) of all entity pairs are calculated through the Relation Feature Calculation module, which is used to construct the original entity-pair matrix ($M$). Thirdly, the Dense-CCNet module transforms $M$ into a context-enhanced entity-pair matrix ($M'$). Finally, the context-enhanced relation features ($M'_{s,o}$) of the entity pairs ($e_s$, $e_o$), the subject entity embedding ($h_{e_s}$), and object entity embedding ($h_{e_o}$) are concatenated and inputted to the classifier to predict the relationship.}
\label{fig2}
\end{figure*}

In this section, we elaborate on our Dense-CCNet mode. Our entire model (as shown in Figure~\ref{fig2}) is mainly composed of three parts: Encoder module (Sec. \ref{sec2.1}), Dense-CCNet module (Sec. \ref{sec2.2}), and Classifier module (Sec. \ref{sec2.3}). 

% \subsection{Task Formulation}

% We first formulate document-level RE task as follows. Given a document composed of $N$ sentences $D\!=\!\{s_i\}_i^N$, which contains $P$ entities $E\!=\!\{e_i\}_i^P$, where $s_i\!=\!\{w_j\}_{j=1}^{n_i}$ refers to the $i{-}th$ sentence in the document consisting of $n_i$ words, $e_i\!=\!\{m_j\}_{j=1}^{p_i}$ represents the set of $p_i$ mentions contained in the $i\!-\!th$ entity, and $m_j$ refers to a span of words belonging to the $j{-}th$ mention, document-level RE aims to extract the relationship between different entities in $E$, namely $\{(e_i,r_{ij},e_j)|e_i,e_j{\in} E,r_{ij}{\in} R\}$, where $R$ is a pre-defined relation type set.
% Then, we define an entity-pair matrix $M{\in}R^{P\times P\times d}$, where $M_{s,o}$ is the correlation feature vector of the entity pair $(e_s,e_o)$ which is related to the relation type between $e_s$ and $e_o$, and $d$ is the dimension of the feature vector.
% In $M$, the entity pair in the  $s{-}th$ row and the $o{-}th$ column overlap the entity with the entity pair $(e_s,e_o)$.

\subsection{Encoder Module}
\label{sec2.1}
We first treat the document $D$ as a sequence of words, i.e. $D\!=\!\{w_i\}_i^L$, where $L$ is the total number of words in the document.
Then, we insert special symbols $\left \langle E_i \right \rangle$ and $\left \langle \setminus E_i \right \rangle$ to mark the start and end positions of the mention respectively, where $E_i$ is the entity id of the mention. 
It is an improved version of entity marker technology \cite{c:112,c:113,c:118,c:103} by introducing entity id information which can help align the information of different mentions from the same entity. 
Finally, we leverage the pre-trained language model as an encoder to convert documents $D$ into a sequence of contextual embeddings as follows:
\begin{equation}
H = [h_1,...,h_L ]= BERT([w_1,...,w_L ]) .
\end{equation}
% Due to the maximum sequence length that BERT can handle is 512, if the length of the document is longer than 512, we make use of a sliding window technology that separates the document into multiple sequences, which follows the paper \cite{c:119}.
We take the embedding of $\left \langle E_i \right \rangle$ at the start of mentions $m_j$ as the mention embedding $h_{m_j}$. Then, we leverage logsumexp pooling \cite{c:120}, a smooth version of max pooling, to obtain the embedding $h_{e_i}$ of entity $e_i$ which contains $p_i$ mentions $\{m_j^i\}_j^{p_i}$:
\begin{equation}
h_{e_i}=\log\sum_{j=1}^{p_i}exp(h_{m_j^i}) .
\end{equation}

After obtaining the embedding of all entities in the document, we construct an \textbf{Entity-Pair Matrix} $M{\in}R^{N_e\times N_e\times d}$ through the Relation Feature Calculation module, where $N_e$ refers to the number of entities and $d$ is the dimension of the relation feature vector. 
The $M_{s,o}$ item in $M$ represents the relation feature vector between the entity $e_s$ and the entity $e_o$, which is calculated as follows:
\begin{equation}
	\begin{split}
	    M_{s,o}&=FFNN([u_s,u_o]) ,   \\
	    u_s&=\tanh(W_s[h_{e_s},h_{doc},c_{s,o}]) , \\
	    u_o&=\tanh(W_o[h_{e_o},h_{doc},c_{s,o}) ,
	\end{split}
\end{equation}
where $h_{e_s}$ is subject entity embedding, $h_{e_o}$ is object entity embedding, $h_{doc}$ is document embedding, and $c_{s,o}$ is entity-pair-aware context feature, $FFN()$ refers to a feed-forward neural network, $W_o,W_s$ is the learnable weight matrix.
% In $M$, the entity pair in the  $s{-}th$ row and the $o{-}th$ column overlap the entity with the entity pair $(e_s,e_o)$.

We use the embedding of the document start token “[CLS]” as the document embedding $h_{doc}$, which can help aggregate cross-sentence information and provide document-aware representation.

The entity-pair-aware context feature $c_{s,o}$ represents the contextual information in the document that the entity $e_s$ and the entity $e_o$ pay attention to together. 
The $c_{s,o}$ is formulated as follows:
\begin{equation}
	\begin{split}
	    c_{s,o}&=\sum_{i=1}^{L} A_{s,i} \cdot A_{o,i} \cdot h_i, \\
	   % Q&=EW_Q^B , K=HW_K^B , V=HW_V^B  ,\\
	   % A&=softmax(\frac{QK^T}{\sqrt{d_k}}) ,\\
	   % S_{ij}&=A_i*A_j ,\quad S_{ij}=S_{ij}/sum(S_{ij}) \\
	\end{split}
\end{equation}
where $A_{s,i}$ is the attention score of the entity $e_s$ paying attention to the i-th token $h_i$ in the document.

% of which calculation process we follow \cite{c:119,c:112}.
% We first extract the multi-headed attention matrix $A{\in}R^{h\times L\times L}$ of the last layer of Bert, where $h$ refers to the number of heads in multi-head self-attention, and apply mean pooling along the attention head dimension (i.e., first dimension) over $A$ to obtain $A'{\in}R^{L\times L}$:
% \begin{equation}
%     A'=\frac{1}{h}\sum_{j}^{h}A_i .
% \end{equation}
% Then, we extract the mention-level attention from $A'$ according to the mention start symbol $〈E_{type} 〉$ and average the attention of mentions to obtain entity-level attention $A^E{\in}R^{N_e\times L}$, which denotes attention from each entity to all tokens in the document.
% Finally, we leverage entity-pair-aware attention vector ($a_{s,o}$) with affine transformation to obtain the entity-pair-aware context feature $c_{s,o}$, as follows:
% \begin{equation}
% 	\begin{split}
% 	    c_{s,o}&=WHa_{s,o} ,   \\
%         a_{s,o}&=softmax(A_s^E*A_o^E) ,
% 	\end{split}
% \end{equation}
% where $W$ is the learnable weight matrix, $*$ refers to element-wise multiplication.

\subsection{Dense-CCNet Module}
\label{sec2.2}
In this part, we introduce the Dense-CCNet module in detail.
As shown in Figure~\ref{fig2}, the Dense-CCNet module consists of densely connected $N$ identical layers that are composed of two sub-modules: the \textbf{Criss-Cross Attention (CCA) module} and the \textbf{Transition module}.

We followed the CCNet model \cite{c:116} for the CCA module.
Each entity pair in the entity-pair matrix can pay attention to the relation feature of other entity pairs in horizontal and vertical directions through the CCA module.
The CCA module can be formulated as follows:
\begin{equation*}
	\begin{split}
	    M_{s,o}&=\sum_{i=1}^{N_e} \Biggl(A_{(s,o)\rightarrow(s,i)} M_{s,i}+A_{(s,o)\rightarrow(i,o)} M_{i,o} \Biggl) \\
	\end{split}
\end{equation*}
where $A_{(s,o)\rightarrow(s,i)}$ is the attention score of the $M_{s,o}$ paying attention to the $M_{s,i}$.
Therefore, the CCA module can complete entity-pair-level one-hop reasoning on the entity-pair matrix, and it is possible to complete multi-hop reasoning by stacking multiple layers of the CCA module.

However, simply using Recurrent Criss-Cross Attention (\textbf{RCCA}) \cite{c:116} to complete the logical reasoning of the entity-pair level may have several problems:
\textbf{(1)} The RCCA only focuses on the high-level multi-hop inference feature and ignores the low-level single-hop inference feature which is also very important for document-level RE.
\textbf{(2)} The CCA module can only model the reasoning mode of \textbf{A$\rightarrow{*}\rightarrow$B}, but cannot model the reasoning mode of \textbf{A$\rightarrow{*}\leftarrow$B},  \textbf{A$\leftarrow{*}\leftarrow$B}, and \textbf{A$\leftarrow{*}\rightarrow$B}.
\textbf{(3)} Since most of the entity pairs are irrelevant in the document, the entity-pair matrix $M$ contains a lot of noise which may affect the reasoning ability of the model.  
Therefore, distinguishing related entity-pairs from unrelated entity-pairs and strengthening the interaction of the relationship feature vectors of related entity-pairs is the key to improving The reasoning ability of the model.
To solve these problems, we have introduced the following methods:

\noindent
\textbf{Dense Connection}: Since dense connections can reuse the features of low-level networks, we stack multiple layers of the CCA modules through the densely connected framework to solve the problem \textbf{(1)} .
In addition, the dense connection can also reduce noise propagation to a certain extent.

\noindent
\textbf{Expanding the Field of Attention}: 
To allow the CCA module to model more inference modes, we modify the CCA module as follows: \begin{equation*}
	\begin{split}
	    M_{s,o}=&\sum_{i=1}^{N_e}\Biggl(A_{(s,o)\rightarrow(s,i)} M_{s,i}+A_{(s,o)\rightarrow(i,o)} M_{i,o} \\
	    &+ A_{(s,o)\rightarrow(s,i)} M_{s,i}+A_{(s,o)\rightarrow(i,o)} M_{i,o} \Biggl) \\
	\end{split}
\end{equation*}
The modified CCA module can cover a wider range of reasoning modes including: \textbf{A$\rightarrow{*}\rightarrow$B}, \textbf{A$\rightarrow{*}\leftarrow$B},  \textbf{A$\leftarrow{*}\leftarrow$B}, and \textbf{A$\leftarrow{*}\rightarrow$B}.

\noindent
\textbf{Clustering Loss}:
We design a clustering loss function that separates the related entity pairs and the unrelated entity pairs in the feature space to reduce the influence of unrelated entity pairs on the inference process. 
Clustering Loss is formulated as follows:
\begin{equation*}
	\begin{split}
	    L_{dist}&=(max\{0, (\mu + cos(\upsilon_0,\upsilon_1))\})^2 ,   \\
	    L_{var1}&=\sum_{i\in N_{pos}}{}(max\{0, (\lambda - cos(\upsilon_1,f_i))\})^2 , \\
	    L_{var0}&=\sum_{j\in N_{neg}}{}(max\{0, (2\lambda - cos(\upsilon_0,f_j))\})^2 , \\
	    L_C&=\alpha  L_{dist} +\beta L_{var0} + \gamma  L_{var1},
	\end{split}
\end{equation*}
where $N_{neg}$ is the set of the irrelevant entity pairs, $N_{pos}$ is the set of the related entity pairs, $\upsilon_0$ is the average vector of the feature vectors of the entity pairs in the $N_{neg}$, $\upsilon_1$ is the average vector of the feature vectors of the entity pairs in the $N_{pos}$, $f_i$ is the feature vector of the i-th entity pair.

\noindent
\textbf{Attention Bias}:
To make the CCA more focused on the related entity pairs, we added a bias to the attention score of the CCA:
\begin{equation}
	\begin{split}
	    A_{(s,o)\rightarrow \star}&=softmax(s_{(s,o)\rightarrow \star}+bias_{\star}) ,   \\
	\end{split}
\end{equation}
where, $bias{\star}$ is a bias term of the entity pair $\star$, which reflects the confidence that the entity pair $\star$ is a related entity pair.
$bias_\star$ is predicted and trained through a feed-forward neural network:
\begin{equation}
	\begin{split}
	    bias_*&=FFNN(f_{\star}),   \\
	    L_{bias}&=BCE(bias_{\star},label_{01})
	\end{split}
\end{equation}
Where $BCE$ is a cross-entropy loss function, and $label_{01}$ is the 0-1 label of the entity pair.

The Transition module controls the dimensions of the new features generated by each layer of the Dense-CCNet model, which reduces the computational complexity of the model.

% In addition, each entity pair is able to harvest full entity-pair matrix contextual information from all entity-pair to generate new correlation features with global interdependence among triples through the stack of two layers of cross-attention modules.
% In order to improve the expressive ability of the model, we use a multi-head mechanism in the Criss-Cross Attention module, i.e., Multi-head Criss-Cross Attention module.

% We stack each layer of the CCNet model through a densely connected framework \cite{c:121} because dense connections can reuse the local interdependence among triples generated by the lower layers of the model.
% With the help of dense connections, we are able to train a deeper model, capture both global and local interdependency among triples, and complete higher-order logical reasoning.

\subsection{Classification Module}
\label{sec2.3}
We use the Dense-CCNet to convert the original entity-pair matrix $M$ into a new context-enhanced entity-pair matrix $M'$.
% which contains richer global and local information.
Given an entity pair $(e_s,e_o)$, we first concatenate the two entity embedding ($h_{e_s}$, $h_{e_o}$) and new relation feature $M'_{so}$, then we obtain the distribution of relationship via a bilinear function. Formally, we have:
\begin{equation}
	\begin{split}
	    z_s&=\tanh(W_s'[h_{e_s},M'_{so}]) , \\
	    z_o&=\tanh(W_o'[h_{e_o},M'_{so}]) ,  \\
	    P(r|e_s,e_o)&=\sigma(z_s^T W_r Z_o +b_r) .
	\end{split}
\end{equation}

For the loss function, we use adaptive-thresholding loss \cite{c:112}, which learns an adaptive threshold for each entity pair. 
% The adaptive-thresholding loss can alleviate the imbalance relation distribution problem (many entity pairs have relation of $NA$) to a certain extent.
The loss function is broken down into two parts as shown below:
\begin{equation*}
	\begin{split}
	    L_1&=-\sum_{r{\in}P_D} \log (\frac{\exp(logit_r)}{\sum_{r'{\in}P_D\cup\{TH\}} exp(logit_{r'})}) ,\\
	    L_2&=-\log (\frac{\exp(logit_{TH})}{\sum_{r'{\in}N_D\cup\{TH\}} exp(logit_{r'})}) ,\\
	    L_{adap}&=L_1+L_2 ,
	\end{split}
\end{equation*}
where $TH$ is an introduced class to separate positive classes and negative classes: positive classes would have higher probabilities than $TH$, and negative classes would have lower probabilities than $TH$, $P_D$ and $N_D$ are the positive classes set and negative classes set in document D  respectively.

Finally, our total loss function is defined as follows:
\begin{equation}
	\begin{split}
	    L&= L_{adap}+L_{bias}+L_{C}, \\
	\end{split}
\end{equation}

\section{Experiments}

\subsection{Datasets}
We evaluate our Dense-CCNet model on three public document-level RE datasets. 
The statistics of the datasets could be found in Appendix~\ref{appendix-a}.
\begin{itemize}
\item \textbf{DocRED} \cite{c:104}:
The DocRED is a large-scale crowdsourced dataset for document-level RE, which was constructed from Wikipedia and Wikidata. 
The DocRED contains 3053 documents for training, 1000 for validating, and 1000 for the test. It involves 97 types of target relations in total, and each document approximately contains 26 entities on average. 
\item \textbf{CDR} \cite{c:122}:
The CDR is a biomedical dataset is constructed by using the PubMed abstracts, which aims to predict the binary interactions between Chemical and Disease concepts. The CDR contains only one relationship and consists of 1500 human-annotated documents in total. 
The CDR are equally split into training, development, and test sets.
\item \textbf{GDA} \cite{c:123}:
Similar to the CDR, the GDA is also a dataset in the biomedical domain, but is constructed by distant supervision from the MEDLINE abstracts. 
The GDA contains 29192 documents as the training set and 1000 as the test set. Since there is no development set, we follow \cite{c:107} to divide the training set into two parts according to the ratio of 8:2 and use them as training set and development set respectively.
\end{itemize}

\begin{table*}[]
\centering
\begin{tabular}{lcccc}
\toprule
Model                & \multicolumn{2}{c}{Dev}   & \multicolumn{2}{c}{Test} \\ 
                     & Ign$F_1$       & $F_1$          & Ign$F_1$      & $F_1$          \\ [2pt] \toprule
GEDA-$\rm BERT_{base}$\cite{c:130}        & 54.52       & 56.16       & 53.71      & 55.74       \\
LSR-$\rm BERT_{base}$ \cite{c:106}        & 52.43       & 59          & 56.97      & 59.05       \\
GLRE-$\rm BERT_{base}$\cite{c:108}        & -           & -           & 55.4       & 57.4        \\
HeterGSAN-$\rm BERT_{base}$\cite{c:134}  & 58.13       & 60.18       & 57.12      & 59.45       \\ 
GAIN-$\rm BERT_{base}$\cite{c:105}       & 59.14       & 61.22       & 59         & 61.24       \\
[2pt] \toprule
$\rm BERT_{base}$\cite{c:135}             & -           & 54.16       & -          & 53.2        \\
BERT-$\rm TS_{base}$\cite{c:135}          & -           & 54.42       & -          & 53.92       \\
HIN-$\rm BERT_{base}$\cite{c:111}         & 54.29       & 56.31       & 53.7       & 55.6        \\
Coref$\rm BERT_{base}$\cite{c:146}        & 55.32       & 57.51       & 54.54      & 56.96       \\
ATLOP-$\rm BERT_{base}$\cite{c:112}       & 59.22       & 61.09       & 59.31      & 61.3        \\
DocuNet-$\rm BERT_{base}$\cite{c:113}     & 59.86       & 61.83       & 59.93      & 61.86       \\ [2pt] \toprule
Dense-CCNet-$\rm BERT_{base}$ & \textbf{60.72($\pm$0.12)} & \textbf{62.74($\pm$0.15)} & \textbf{60.46} & \textbf{62.55} \\ \bottomrule
% w/o Dense                               & 59.56(-1.21)       & 61.65(-0.82)   & -      & - \\
% w/o Dense-CCNet                         & 58.93(-1.84)       & 60.92(-1.55)   & -      & -   \\ 

\end{tabular}
\caption{\label{tab1} Results (\%) on the development and test set of the DocRED. We follow ATLOP \cite{c:112} and DocuNet \cite{c:113} for the scores of all baseline models. The results on the test set are obtained by submitting to the official Codalab.
% w/o Dense means using residual connections to replace dense connections. w/o Dense-CCNet means to remove the Dense-CCNet module from our entire model.
}
\end{table*}

\subsection{Experimental Settings}
Our model was implemented based on PyTorch. We used cased BERT-base \cite{c:125} as the encoder on DocRED and SciBERT-base \cite{c:127} on CDR and GDA.
We set the number of layers of Dense-CCNet to 3.
Our model is optimized with AdamW \cite{c:128} with a linear warmup \cite{c:129} for the first 6\% steps followed by a linear decay to 0. 
All hyper-parameters are tuned on the development set, some of which are listed in Appendix~\ref{appendix-b}.

\subsection{Results on the DocRED Dataset}
The experimental results of our model on DocRED are shown in Table~\ref{tab1}. We followed \cite{c:104} and used $F_1$ and Ign$F_1$ as the evaluation metrics to evaluate the overall performance of the model. Ign$F_1$ denotes the $F1$ score excluding the relational facts that are shared by the training and dev/test sets. We compare the Dense-CCNet model with the following two types of models on the DocRED dataset :
\begin{itemize}
    \item \textbf{Graph-based Models}: these models first construct the document-graph from the document, and then perform inferences through GCN \cite{c:110} on the graph. We include GEDA \cite{c:130}, LSR \cite{c:106}, GLRE \cite{c:108}, GAIN \cite{c:105}, and HeterGSAN \cite{c:134} for comparison.
    
    \item \textbf{Transformer-based Models}: These models directly use the pre-trained language models for document-level RE without graph structures. we compared $\rm BERT_{base}$ \cite{c:135}, BERT-$\rm TS_{base}$ \cite{c:135}, HIN-$\rm BERT_{base}$ \cite{c:111}, $\rm CorefBERT_{base}$ \cite{c:146}, Coref$\rm BERT_{base}$\cite{c:146}, and ATLOP-$\rm BERT_{base}$ \cite{c:112} with our model.
\end{itemize}
In addition, we also consider the DocuNet \cite{c:113} model in the comparison, which formulates document-level RE as a semantic segmentation problem. 
% The DocuNet model is the state-of-the-art model recently proposed.
% and captures global interdependency among triples through the U-shaped segmentation model \cite{c:115}.

As shown in Table~\ref{tab1}, our Dense-CCNet model achieved \textbf{62.74}\%$F_1$  and \textbf{62.55}\% $F_1$ in the training set and test set, which outperforms the state-of-the-art model with \textbf{0.91}\% $F_1$ and \textbf{0.69}\% $F_1$ respectively.
% Our Dense-CCNet model outperforms the DocuNet model by \textbf{0.91}\% Ign$F_1$ on the dev set and \textbf{0.27}\% Ign$F_1$ on the test set, which shows that our model can capture the interdependence among triples more effectively than the DocuNet.
Compared with the GAIN model that is the state-of-the-art model of graph-based methods, our model exceeds it by \textbf{1.52}\% $F_1$ on the dev set and \textbf{1.31}\% $F_1$ on the test set.
This proves that the logical reasoning on the entity-pairs level is more effective than previous methods on mentions or entities level.

\subsection{Results on the Biomedical Datasets}
% On the CDR and GDA data sets, we found that the relationship of many entity pairs is directly represented by their mention pairs. Therefore, we extend our method to mention-level to obtain better performance.
% Specifically, we first construct a mention-pair matrix in the same way as the entity-pair matrix and then apply the Dense-CCNet module to the matrix to obtain a context-enhanced mention-pair matrix. In the end, I apply the average pooling operation on the context-enhanced mention-pair matrix to get the mention-wise entity pair representation $m_{s,o}$.
% To integrate $m_{s,o}$ into the entity-pair matrix, we modify the calculation process of entity-pair matrix $M$ (Formula (3)) as follows:
% \begin{equation}
% 	\begin{split}
% 	    M_{s,o}&=FFN([u_s,u_o])   ,\\
% 	    u_s&=\tanh(W_s[h_{e_s},h_{doc},c_{s,o},m_{s,o}]) ,\\
% 	    u_o&=\tanh(W_o[h_{e_o},h_{doc},c_{s,o},m_{s,o}]) .
% 	\end{split}
% \end{equation}

% Due to the small size of the CDR dataset, we select to merge the train and development sets and re-train our model on the union for evaluation on the test set following \cite{c:153,c:107,c:108}. 
On the CDR and GDA data sets, we compared BRAN \cite{c:137}, EoG \cite{c:107}, LSR \cite{c:106}, DHG \cite{c:139}, GLRE \cite{c:108}, ATLOP \cite{c:112}, and DocuNet \cite{c:113} with our model.
The experimental results on two biomedical datasets are shown in Table~\ref{tab2}.

Our Dense-CCNet-$\rm SciBERT_{base}$ model obtained \textbf{77.06($\pm$0.71)}\% $F_1$ and \textbf{86.44($\pm$0.25)}\% $F_1$ on two data sets respectively, which is also the new state-of-the-art result.
The Dense-CCNet-$\rm SciBERT_{base}$  improved the $F_1$ score by \textbf{0.76}\% and \textbf{1.14}\% on CDR and GDA compared with DocuNet-$\rm SciBERT_{base}$. 
These results demonstrate the strong applicability and generality of our approach in the biomedical field.

% \begin{table}[]
% \centering
% \setlength{\tabcolsep}{4.5mm}{
% \begin{tabular}{lcc}
% \toprule
% Model                                   & \multicolumn{2}{c}{Dev}   \\ 
%                                         & Ign$F_1$      & $F_1$     \\ [2pt] \toprule
% Dense-CCNet-$\rm BERT_{base}$  & \textbf{60.77} & \textbf{62.47}         \\ 
% w/o Dense                               & 59.56       & 61.65       \\
% w/o Dense-CCNet                         & 58.93       & 60.92       \\ \bottomrule

% \end{tabular}}
% \caption{\label{font-table} Ablation study of Dense-CCNet on the development set of DocRE. w/o Dens means using residual connections to replace dense connections. w/o Dense-CCNet means to remove the Dense-CCNet module from our entire model.}
% \end{table}

\begin{table}[]
\centering
\begin{tabular}{p{4.7cm}cc} 
\toprule
Model               & CDR          & GDA         \\ [2pt] \toprule
BRAN \cite{c:137}                & 62.1         & -           \\
EoG \cite{c:107}                & 63.6         & 81.5        \\
LSR \cite{c:106}                & 64.8         & 82.2        \\
DHG \cite{c:139}                & 65.9         & 83.1        \\
GLRE \cite{c:108}                & 68.5         & -           \\
$\rm SciBERT_{base}$ \cite{c:127}        & 65.1         & 82.5        \\
ATLOP-$\rm SciBERT_{base}$\cite{c:112}   & 69.4         & 83.9        \\
DocuNet-$\rm SciBERT_{base}$\cite{c:113} & 76.3         & 85.3        \\  [2pt] \toprule
Dense-CCNet-$\rm SciBERT_{base}$         & \textbf{77.06} & \textbf{86.44}    \\
\bottomrule
\end{tabular}
\caption{\label{tab2} F1 scores (\%) on test sets of the CDR and the GDA.}
\end{table}

\subsection{Ablation Study}
We conducted an ablation experiment to validate the effectiveness of different components of our Dense-CCNet model on the development set of the DocRED dataset.
The results are listed in Table~\ref{tab3}, where 
\textbf{w/o Dense Connection} replaces the Densely connected Criss-Cross Attention with the Recurrent Criss-Cross Attention (RCCA),
\textbf{w/o Expanding Attention} uses standard Criss-Cross Attention and does not extend the field of attention, 
\textbf{w/o Clustering Loss} and \textbf{w/o Attention Bias} respectively removes the Clustering Loss and the Attention Bias from our model.

From Table~\ref{tab3}, we can observe that the w/o Dense leads to a drop of \textbf{1.62\%} $F1$, which shows that the features of low-level inference are very helpful for relation extraction and  the features of high-level inference may contain noises. 
The w/o Expanding Attention caused a performance drop of \textbf{0.83\%} $F1$, which indicates that document-level RE may include multiple inference modes and our model can effectively expand the reasoning mode of the Criss-Cross Attention through the Expanding the Field of Attention technology.

The w/o Clustering Loss module and w/o Attention Bias module led to performance degradation of \textbf{0.72\%} $F1$ and \textbf{1.14\%} $F1$ points respectively, which reflects that reducing noise (irrelevant entity-pairs) may be the key to further improving entity-pair level inference. 
We guess that the most ideal entity-pair-level reasoning method may be to only propagate information between related entity pairs.

In addition, we also introduced the ablation study of the number of layers of the Dense-CCNet, and the experimental results are shown in Table~\ref{tab4}.
When the number of layers increases from 2 to 3, our model can capture more multi-hop inference features, so the performance of the model is improved by \textbf{1.3\%} $F1$.
However, when the number of layers is increased to 4, the performance drops slightly by \textbf{0.37\%} $F1$ points.
The possible reasons is that the noise has a greater impact on the high-level feature or the model falls into over-fitting.

\subsection{Case Study}
We followed GAIN \cite{c:105} to select the same example and conduct a case study to further illustrate that our Dense-CCNet model can effectively capture the interdependence among entity-pairs and perform entity-pair-level logical reasoning compared with the baseline.

\begin{figure*}[t]
\centering
\includegraphics[width=2.0 \columnwidth]{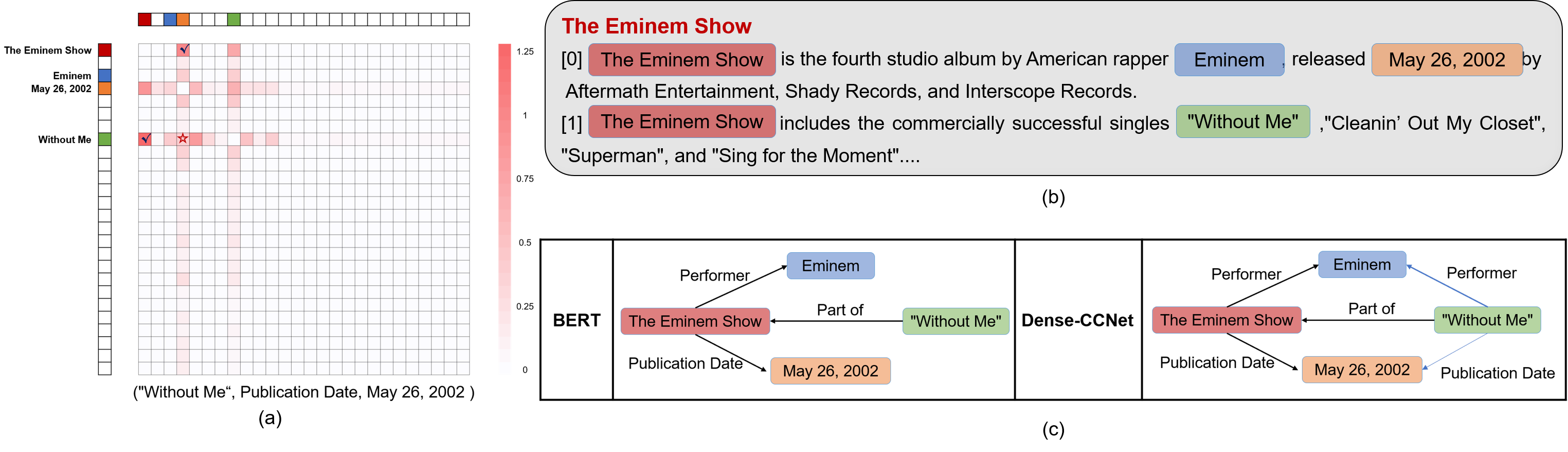} % Reduce the figure size so that it is slightly narrower than the column. Don't use precise values for figure width.This setup will avoid overfull boxes.
\caption{Case study of our Dense-CCNet mode and the baseline model. (c) shows that our model has better logical reasoning ability than the baseline. (a) visualize the attention scores of entity pairs (\textit{``Without Me", May 26, 2002}) paying attention to other entity pairs, which shows that our model can effectively capture the correlation among entity-pairs.}
\label{fig3}
\end{figure*}

The experimental results are shown in Figure~\ref{fig3}. Figure 3c demonstrates that our model has better logical reasoning ability than the baseline.
Figure 3a shows that the entity pair (\textit{``Without Me", May 26, 2002}) has more attention to the entity pairs (\textit{``Without Me", The Eminem Show}) and (\textit{The Eminem Show, May 26,2002}),  which indicates that our model could capture the correlation these among entity-pairs.

\begin{table}[]
\centering
\setlength{\tabcolsep}{2.8mm}{
\begin{tabular}{lcc}
\toprule
Model                                   & \multicolumn{2}{c}{Dev}   \\ 
                                        & Ign$F_1$      & $F_1$     \\ [2pt] \toprule
Dense-CCNet-BERT                     & \textbf{60.72}   & \textbf{62.74}         \\ 
w/o Dense Connection                       & 59.23       & 61.12       \\
w/o Expanding Attention                      & 59.82       & 61.91       \\
w/o Clustering Loss                      & 59.97       & 62.02        \\
w/o Attention Bias                & 60.65       & 61.60        \\   \bottomrule
\end{tabular}}
\caption{\label{tab3} Ablation study of the Dense-CCNet on the development set of the DocRED. We turn off different components of the model one at a time.
}
\end{table}

\begin{table}[]
\centering
\setlength{\tabcolsep}{5.5mm}{
\begin{tabular}{lcc}
\toprule
Layer-number                           & \multicolumn{2}{c}{Dev}   \\ 
                                        & Ign$F_1$      & $F_1$     \\ [2pt] \toprule
2-Layer                           & 59.41       & 61.44       \\
3-Layer                          & \textbf{60.72} & \textbf{62.74}       \\
4-Layer                           & 60.30       & 62.27       \\ \bottomrule
% 6-Layer                     & 59.26       & 61.36       \\ \bottomrule

\end{tabular}}
\caption{\label{tab4} Performance of the Dense-CCNet with the different numbers of layers on the development set of the DocRE.}
\end{table}

\section{Related Work}
\textbf{Sentence-level RE}: 
Early research on RE focused on sentence-level RE, which predicts the relationship between two entities in a single sentence.
Many approaches \cite{c:140,c:141,c:143,c:144,c:145,c:146,c:147,c:148,c:149,c:150} have been proven to effectively solve this problem.
Since many relational facts in real applications can only be recognized across sentences, sentence-level RE face an inevitable restriction in practice.

\noindent \textbf{Document-level RE}: 
To solve the limitations of sentence-level RE in reality, a lot of recent work gradually shift their attention to document-level RE.
Since graph neural network(GNN) can effectively model long-distance dependence and complete logical reasoning, Many methods based on document-graphs are widely used for document-level RE.
Specifically, they first constructed a graph structure from the document, and then applied the GCN \cite{c:110,c:121} to the graph to complete logical reasoning.
The graph-based method was first introduced by \cite{c:152} and has recently been extended by many works \cite{c:107,c:130,c:139,c:151,c:108,c:106,c:105,c:134}.
\cite{c:130} proposed the Graph Enhanced Dual Attention network (GEDA) model and used it to characterize the complex interaction between sentences and potential relation instances.
\cite{c:105} propose Graph Aggregation-and-Inference Network (GAIN) model.
GAIN first constructs a heterogeneous mention-level graph (hMG) to model complex interaction among different mentions across the document and then constructs an entity-level graph (EG), finally uses the path reasoning mechanism to infer relations between entities on EG.
\cite{c:106} proposed a novel LSR model, which constructs a latent document-level graph and completes logical reasoning on the graph.

In addition, due to the pre-trained language model based on the transformer architecture can implicitly model long-distance dependence and complete logical reasoning, some studies \cite{c:111,c:112,c:135} directly apply pre-trained model without introducing document graphs.
\cite{c:112} proposed an ATLOP model that consists of two parts: adaptive thresholding and localized context pooling, to solve the multi-label and multi-entity problems.
Recently, the state-of-the-ar model, DocuNet \cite{c:113},  formulates document-level RE as semantic segmentation task and capture  global information among relational triples through the U-shaped segmentation module \cite{c:115}.

% Both types of methods have two potential problems: (1) they mainly focused on the local entity representation, but ignore the global interdependence among triples. (2) They perform logical reasoning only on mentions or entities level rather than entity-pairs level. 
% To capture the interdependency among triples, DocuNet \cite{c:113} formulates document-level RE as semantic segmentation task and capture  global information among relational triples through the U-shaped segmentation module \cite{c:115}.

However, none of the models completes the logical reasoning for document-level RE through the information propagation between the entity-pairs.
Our Dense-CCNet model can capture the correlation among entity-pairs and complete the entity-pair-level reasoning by integrating the CCA \cite{c:116} into the dense connection framework \cite{c:121}.

\section{Conclusion and Future Work}
In this work, we propose a novel Dense-CCNet model by integrating the Criss-Cross Attention into the densely connected framework. 
Dense-CCNet model can complete entity-pairs-level logical reasoning and model the correlation between entity pairs.
Experiments on three public document-level RE datasets demonstrate that our Dense-CCNet model achieved better results than the existing state-of-the-art model.
In the future, we will try to use our model for other inter-sentence or document-level tasks, such as cross-sentence collective event detection.

% Entries for the entire Anthology, followed by custom entries
\bibliography{anthology,custom,acl_latex.bbl}
\bibliographystyle{acl_natbib}
\begin{table*}[]
\centering
\setlength{\tabcolsep}{7.0mm}{
\begin{tabular}{lccc}
\toprule
Dataset                     & DocRED     &CDR       &GDA  \\  [2pt] \toprule
Train                     &3053	    &500	    &23353         \\ 
Dev                       &1000	    &500	    &5839       \\
Test                      &1000	    &500	    &1000       \\
Relations                 &97	    &2	    &2        \\
Entities per Doc        &19.5	    &7.6	    &5.4        \\
Mentions per Doc        &26.2	    &19.2	    &18.5        \\ 
Entities per Sent        &3.58	    &2.48	    &2.28        \\     \bottomrule

\end{tabular}}
\caption{\label{tab5} Summary of DocRED, CDR and GDA datasets.
%The ablation results show that every components of our model is effective.
}
\end{table*}

\begin{table*}[]
\centering
\setlength{\tabcolsep}{7.0mm}{
\begin{tabular}{lccc}
\toprule
Hyperparam                     & DocRED     &CDR       &GDA  \\ 
                            & BERT     &SciBERT       &SciBERT  \\  [2pt] \toprule
Batch size                    &8	    &16	    &16         \\ 
Epoch                       &100	    &20	    &5 \\
lr for encoder                 &2e-5	    &1e-5	    &1e-5        \\
lr for other parts       &1e-4	    &5e-5	    &5e-5         \\
\{$\mu$, $\lambda$\}                 &\{1, 0.5\}	    &\{1, 0.5\}	    &\{1, 0.5\}        \\
\{$\alpha$, $\beta$, $\gamma$\}      &\{1, 1, 1\}	   &\{1, 1, 1\}    &\{1, 1, 1\}        \\\bottomrule

\end{tabular}}
\caption{\label{tab6} Hyper-parameters Setting.}
\end{table*}

\appendix
\newpage
\section{Datasets}
\label{appendix-a}
Table~\ref{tab5} details the statistics of the three document-level relational extraction datasets, DocRED, CDR, and GDA. These statis-tics further demonstrate the complexity of entity structure in document-level relation extraction tasks.

\section{Hyper-parameters Setting}
\label{appendix-b}
Table~\ref{tab6} details our hyper-parameters setting. All of our hyperparameters were tuned on the development set.

\end{document}